\documentclass[sigconf]{acmart}

\usepackage{booktabs} 

\setcopyright{none}

\begin{document}

\copyrightyear{2019} 
\acmYear{2019} 
\setcopyright{acmcopyright}
\acmConference[NICE '19]{Neuro-Inspired Computational Elements Workshop}{March 26--28, 2019}{Albany, NY, USA}
\acmBooktitle{NICE '19: Neuro-Inspired Computational Elements Workshop, March 26--28, 2019, Albany, NY, USA}
\acmPrice{15.00}
\acmDOI{0000}
\acmISBN{0000}

\title{Introducing Astrocytes on a Neuromorphic Processor: Synchronization, Local Plasticity and Edge of Chaos}

\author{Guangzhi Tang}
\affiliation{%
  \institution{Computational Brain Lab\\Rutgers University}
  \city{New Brunswick}
  \state{New Jersey}
  \postcode{08854}
}
\email{gt235@cs.rutgers.edu}

\author{Ioannis E. Polykretis}
\affiliation{%
  \institution{Computational Brain Lab\\Rutgers University}
  \city{New Brunswick}
  \state{New Jersey}
  \postcode{08854}
}
\email{ip211@cs.rutgers.edu}

\author{Vladimir A. Ivanov}
\affiliation{%
  \institution{Computational Brain Lab\\Rutgers University}
  \city{New Brunswick}
  \state{New Jersey}
  \postcode{08854}
}
\email{vai9@cs.rutgers.edu}

\author{Arpit Shah}
\affiliation{%
  \institution{Computational Brain Lab\\Rutgers University}
  \city{New Brunswick}
  \state{New Jersey}
  \postcode{08854}
}
\email{arpit.shah@rutgers.edu}

\author{Konstantinos P. Michmizos}
\affiliation{%
  \institution{Computational Brain Lab\\Rutgers University}
  \city{New Brunswick}
  \state{New Jersey}
  \postcode{08854}
}
\email{michmizos@cs.rutgers.edu}

\renewcommand{\shortauthors}{G. Tang et al.}
\renewcommand{\shorttitle}{Introducing Astrocytes on a Neuromorphic Processor}

\begin{abstract}
While there is still a lot to learn about astrocytes and their neuromodulatory role in the spatial and temporal integration of neuronal activity, their introduction to neuromorphic hardware is timely, facilitating their computational exploration in basic science questions as well as their exploitation in real-world applications. Here, we present an astrocytic module that enables the development of a spiking Neuronal-Astrocytic Network (SNAN) into Intel's Loihi neuromorphic chip. The basis of the Loihi module is an end-to-end biophysically plausible compartmental model of an astrocyte that simulates the intracellular activity in response to the synaptic activity in space and time. To demonstrate the functional role of astrocytes in SNAN, we describe how an astrocyte may sense and induce activity-dependent neuronal synchronization, switch on and off spike-time-dependent plasticity (STDP) to introduce single-shot learning, and monitor the transition between ordered and chaotic activity at the synaptic space. Our module may serve as an extension for neuromorphic hardware, by either replicating or exploring the distinct computational roles that astrocytes have in forming biological intelligence.
\end{abstract}
%



\maketitle

\section{Introduction}

Shadowed by a century of neuronal recordings, astrocytes, the electrically silent non-neuronal cells in the brain, have remained absent from most studies on the biological principles of intelligence and from efforts to translate this understanding to computational primitives of artificial intelligence. Recent advancements in $Ca^{2+}$ imaging \cite{volterra2014astrocyte} and selective stimulation \cite{pereaoptogenetic} have signified the active signaling that occurs among astrocytes as well as between astrocytes and neurons \cite{volterra2005}. Astrocytes and other glial cells are now known to do far more than just providing nutritional and structural support to neurons \cite{volterra2005}. Among the many roles attributed to astrocytes in brain function and dysfunction \cite{barres2008}, a striking one is their neuromodulatory ability that originates at the cellular level, where synaptic plasticity \cite{HaydonCSH, Arlington} and neuronal synchronization \cite{Fellin, Knoxville} take place, and extends to the network level, where brain rhythms are observed \cite{LeoRhythms, fries2015}, which is a major component of behavioral functions including memory \cite{LeoMemory, adamsky2018astrocytic} and cognition \cite{Perea2014, han2013, fries2015}.

Astrocytic neuromodulation takes place in several brain regions \cite{araque2014gliotransmitters, Letellier, pereaoptogenetic} and has elevated the synapse to a "tripartite" unit (Fig. 1A), where neurons and astrocytes process and learn information independently and collaboratively \cite{perea2009tripartite}. The current hypothesis is that, by forming a large number of tripartite synapses, an astrocyte acts as a spatiotemporal integrator of the synaptic activity \cite{deemyad2018}. We have recently developed computational models of astrocytic cells that replicate their experimentally reported intra- and inter-cellular activity. Briefly, our models suggest intracellular mechanisms that astrocytes may use to encode and modulate both the excitatory \cite{Arlington, Knoxville} and the inhibitory \cite{Chicago} synaptic activity. The astrocytic models sense the neurotransmitters' spillover from the presynaptic neurons and release gliotransmitters to the postsynaptic neurons, suggesting a spatially constrained cascade mechanism for neuromodulation. Upon binding to astrocytic receptors, the presynaptic neurotransmitters produce an intracellular messenger molecule, Inositol 1,4,5-Triphosphate ($IP_3$); When the $IP_3$ concentration exceeds a threshold, an intracellular $Ca^{2+}$ wave is generated, having a similar temporal and spatial profile to the one observed in \textit{in-vitro} studies \cite{shigetomi2016probing}. This $Ca^{2+}$ wave triggers the release of gliotransmitters which in turn induce a slow inward current ($I_{SIC}$) into the postsynaptic neuron \cite{parpura2000physiological}.

Following the prevailing dogma that brain computation equals neuronal processing, neuromorphic processors are built on the premise that neurons are the \textit{only} computing unit. To overcome the intrinsic limitations of the connectionist models of neuronal computation, multiple neuromorphic chips have now become available \cite{davies2018loihi,merolla2014million,schemmel2010wafer,furber2014spinnaker}. These chips employ a time-dependent computational formalism, spiking neural networks (SNN), where asynchronous computing units are simulated as spiking neurons and memory is distributed at the neuronal synapses. Notably, Intel's Loihi processor is the most advanced large-scale neuromorphic hardware, offering multi-compartmental processing, on-chip learning and an unconstrained scalability \cite{davies2018loihi}. A handful of Loihi applications have already demonstrated unparalleled power efficiency in batch-mode image processing \cite{lin2018programming,davies2018loihi} and speech recognition \cite{blouw2018benchmarking} as well as in real-time control of a mobile robot, demonstrated by our Lab \cite{tang2019spiking}. 

Alongside efforts to implement backprop algorithms in SNN \cite{shrestha2018slayer}, most neuromorphic algorithms are either restricted enough to be dictated by the underlying connectome associated with the targeted function \cite{tang2019spiking}, or simple enough to be trained via variations of spike-time-dependent plasticity (STDP), a biologically relevant yet computationally weak Hebbian-type rule \cite{diehl2015unsupervised,bing2018end}. By correlating pre- and post-synaptic activities, STDP rules contribute to the asynchronous parallelism of neuromorphic chips but they rather oversimplify other types of global, localized or temporal learning rules in the brain \cite{Fitzpatrick}. For example, STDP considers the synaptic activity solely within a single synapse and within a milliseconds-scale, contrary to what is happening in the brain. This inhibits SNN from taking place in regional or global tasks such as synchronization, a well-studied yet poorly-understood brain principle believed to convey information in space and time \cite{fries2015}. Further, by learning short-term correlations, STDP ignores events at larger time-scales, which are crucial for real-time applications. The asynchronous neuromorphic chips lack an activity-dependent mechanism to entrain their SNN into rhythmic activities, much like the brain does to efficiently and timely compute. Also, SNN inherited from the ANN the assumption that learning takes place only in the connection strengths between neurons, although structural plasticity rules that span beyond the single synapse have been known for many years to increase the computational capacity of neurons \cite{poirazi2001}. In fact, learning and memory are optimally balanced in networks that macroscopically operate at the edge of chaos \cite{legenstein2007edge}, a narrow dynamical regime largely ignored in learning algorithms, yet also exhibited by the brain \cite{friedman2012universal}. Overall, neuromorphic solutions seem to be missing a \textit{mesoscopic} learning mechanism that can combine the computational efficiency of having a global network goal with the versatility of a local, activity-dependent, plastic mechanism.

In this paper, we introduce spiking Neuronal-Astrocytic Networks (SNAN) and demonstrate some of the distinct computational and learning abilitities that astrocytes may add to neuromorphic hardware. We first describe our Loihi module\footnote{The module is available at https://github.com/combra-lab/combra\_loihi}, where astrocytes are used as an information processing unit, communicating with neurons in the tripartite synapses and with each other via intra- and inter-cellular $Ca^{2+}$ waves \cite{scemes2006astrocyte}. We then exhibit three example uses of our Loihi module, where astrocytes can extend learning at the meso-scale: Specifically, a) we show how astrocytes may modulate the neuronal component of the SNAN, by using their $Ca^{2+}$ dependent SIC to impose an on-demand synchronization across neuronal areas; b) We present how they can be used for single-shot pattern memorization via astrocyte-reinforced STDP; c) We finally show how a single astrocytic cell may be used to continuously monitor the neuronal component for its transition from order to chaos. With SNAN already showing their potential in a growing number of applications as reinforcement to neuronal networks \cite{porto2011artificial,valenza2013novel,LeoMemory}, the proposed Loihi module can become a computational framework for exploring and exploiting the unique computational principles that astrocytes are now known to exhibit.

\section{The Loihi Astrocytic Module (LAM)}

\begin{figure}
\includegraphics[scale=1]{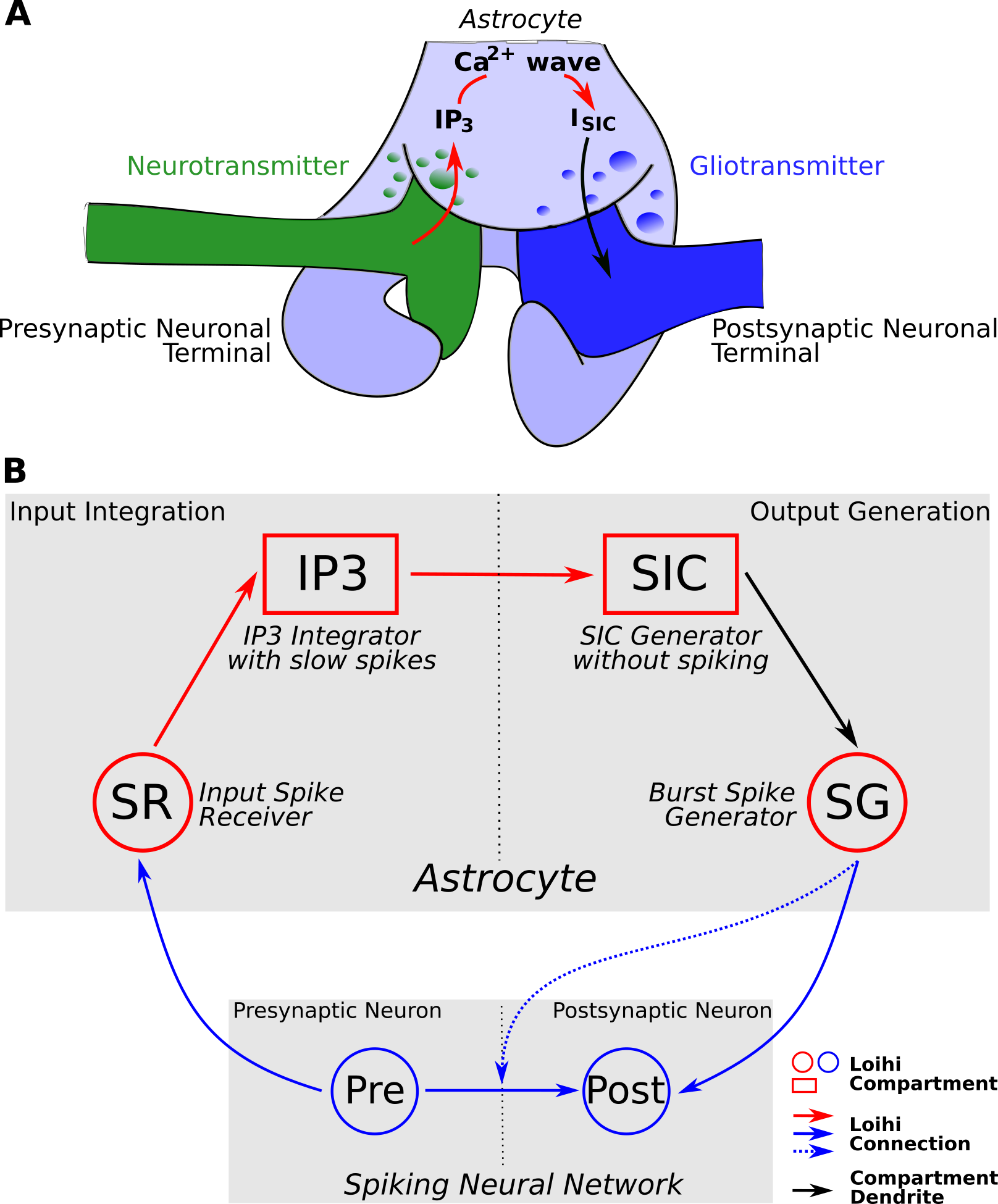}
\caption{A. Schematic of the biological tripartite synapse. B. Its implementation on our Loihi Astrocytic Module, preserving the communication pathways between the astrocytic and the neuronal components of the SNAN.}
\end{figure}

\subsection{Module Overview}

\begin{figure*}
\includegraphics[scale=1]{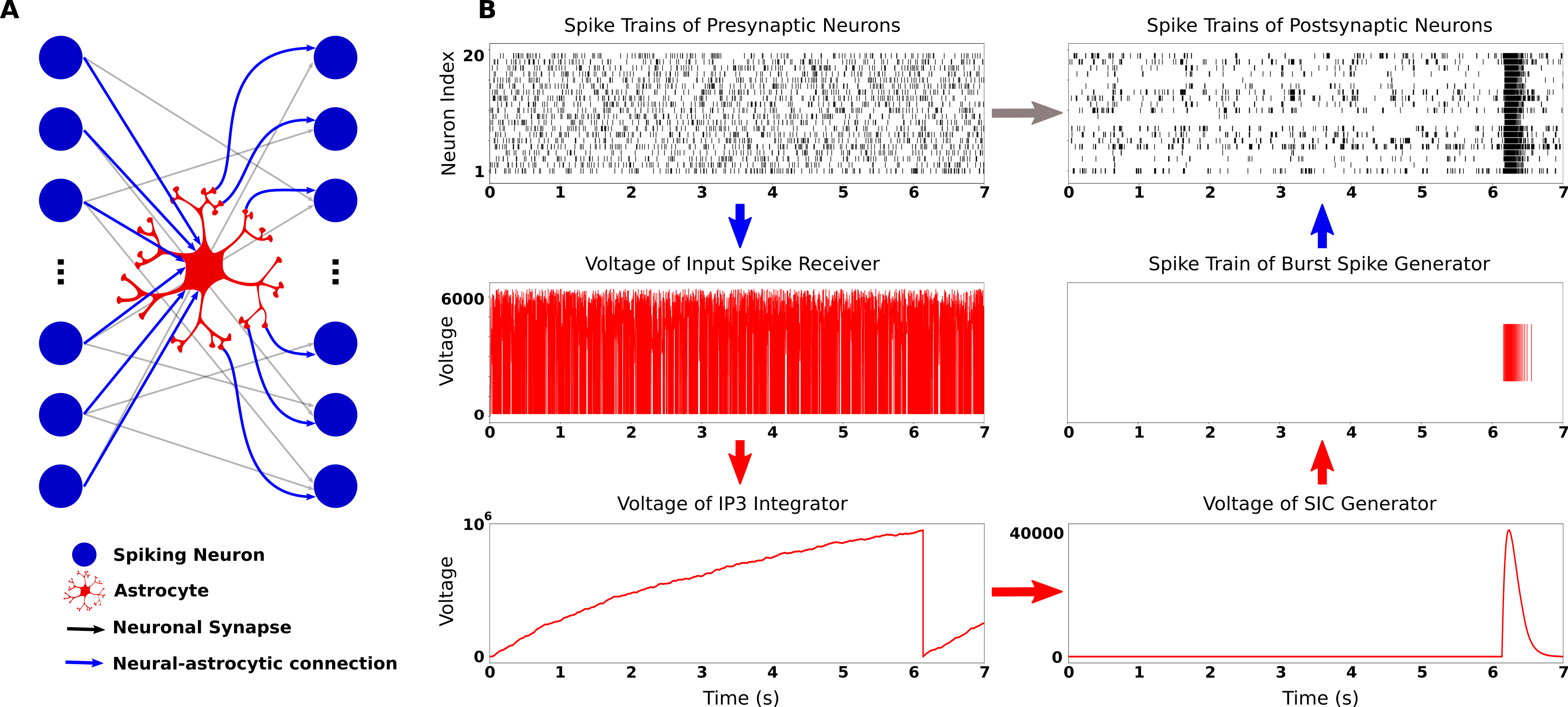}
\caption{A feedforward SNAN where an astrocyte senses and imposes synchronous neuronal activity. A. SNAN Architecture. B. Activity of the astrocyte's main compartments: Integrating synaptic activity induces a SIC injected to postsynaptic neurons.}
\end{figure*}

Our Loihi Astrocytic Module (LAM) emulates the basic communication pathways between neurons and astrocytes in the tripartite synapse as well as the sub-cellular processes that astrocytes use to generate calcium waves in response to synaptic activity. The module's building block is the astrocytic cell that is created as an assembly of 4 Loihi compartments that approximate in hardware the leaky-integrate-and-fire neuronal model. We connected these compartments and tweaked their internal dynamics to represent the slower astrocytic processes (Fig. 1B). This ensures that the astrocytic component can be seamlessly integrated into any spiking neural network (SNN), by only declaring the connections between any of its spiking neurons and astrocytes. 

\subsection{Module Implementation}
The LAM has 2 main components: (i) The input integration component that senses the neuronal activity by emulating the astrocytic receptors and $IP_3$ production, and (ii) the output generation component that modulates the neuronal component of the SNAN. The input integration component is comprised of 2 spiking Loihi compartments, the input Spike Receiver (SR) and the $IP_3$ integrator (IP3). The SR compartment integrates synaptic activity into IP3, resembling the behavior of astrocytic receptors. The IP3 dynamics are much slower than the neuronal dynamics, with their time-scale ranging from hundreds of milliseconds to tens of seconds. This allows for a temporal integration of the synaptic activity. When IP3 levels reach a threshold, the output component generates bursting spikes to either the postsynaptic neurons or the Loihi's reinforcement channel of neuronal connections. The output generation component has a dendritic tree where the non-spiking SIC generator compartment (SIC) feeds its voltage into the burst Spike Generator compartment (SG). The SIC produces a continuous voltage signal similar in shape to the biological $I_{SIC}$. The SG discretizes the signal into bursting spikes sent to the user-defined destinations. 

\subsection{Developing a feedforward SNAN using the LAM}
To demonstrate the internal dynamics of the astrocyte compartments as the mechanisms that an astrocyte uses to impose synchronous activity in postsynaptic neurons \cite{Fellin}, we present a simple 2-layer feedforward SNAN (Fig. 2A). In this network, presynaptic neurons with Poisson spike trains were randomly connected to the same number of postsynaptic neurons. The astrocyte integrated the presynaptic activity and generated an IP3 spike at 6 seconds, which, in turn, triggered the SG bursting spikes and generated synchronous activity in the postsynaptic neurons for 400 milliseconds (Fig. 2B). The astrocytic dynamics are configurable in our astrocyte SDK that is described below.

\section{Astrocyte SDK}

\begin{figure}[b]
\includegraphics[scale=1]{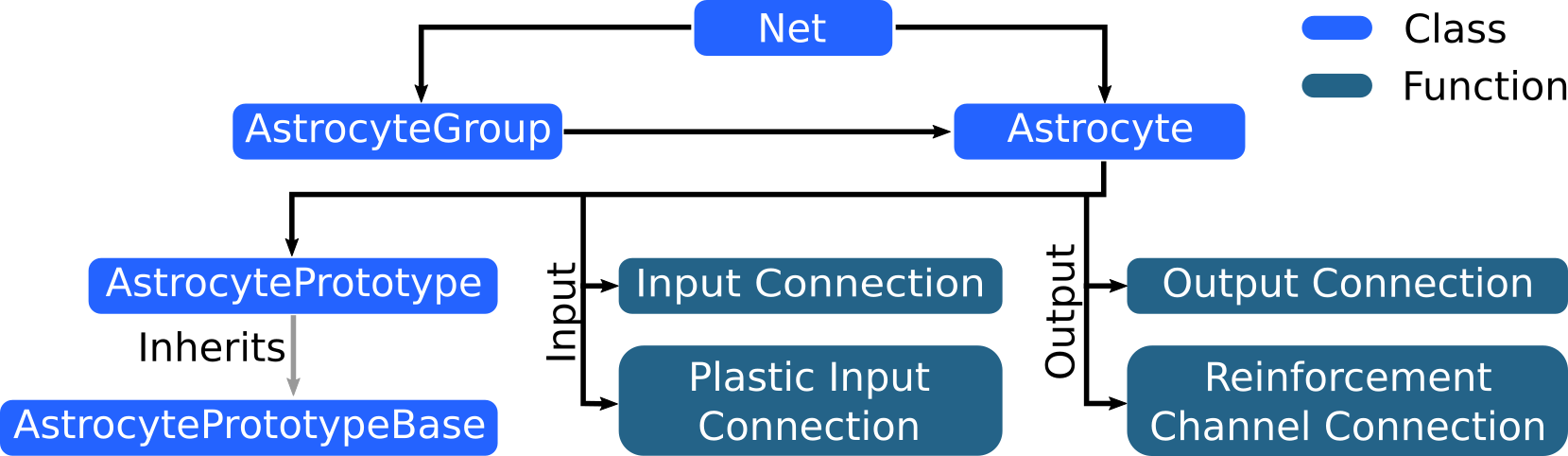}
\caption{Overall Astrocyte SDK Structure.}
\end{figure}

\subsection{SDK Architecture}

The Astrocyte SDK builds on the existing NxSDK Python-based programming model for Loihi \cite{lin2018programming} and it is designed to facilitate the integration of astrocytes with SNNs. The SDK's API provides access to 3 main classes, depicted in Figure 3:
\begin{enumerate}
  \item \textbf{AstrocytePrototype}: The prototype class for defining astrocytes allows users to configure the properties of each of the four compartments within the astrocyte, including the spiking thresholds for the IP3 compartment, the current decay of the SIC compartment and the weights between the compartments. 
  \item \textbf{Astrocyte}: An astrocyte instance describes the internal compartments of the astrocyte model. This class provides the functions for specifying the presynaptic (input) neurons and the postsynaptic (output) neurons, to create tripartite synapses. Each instance of the Astrocyte class requires an AstrocytePrototype; If none is provided, a default prototype is created.
  \item \textbf{AstrocyteGroup}: A wrapper class that instantiates multiple astrocytes. The constructor accepts lists of prototypes, the size of the group of astrocytes and other configurations for mapping astrocyte instances to prototypes and Loihi's logical cores.
\end{enumerate}

\subsection{Automatic Parameter Setup}

Our Astrocyte SDK allows the user to define astrocyte behavior by controlling the following three parameters in the AstrocytePrototype class:

\begin{enumerate}
    \item \textbf{ip3\_sensitivity}: the weight between the SR and the IP3.
    \item \textbf{sic\_amplitude}: the maximum firing rate of the SG.
    \item \textbf{sic\_window}: time duration in milliseconds between the first and last of the spikes generated by the SG.
\end{enumerate}

These SDK level parameters are automatically mapped to Loihi parameters using a precalculated configuration table, generated using the $sic\_data\_table.py$ file located within the utils folder of the module. This script performs a brute force search of Loihi parameter configurations, saving all configurations that yield a minimal discretized SIC output, or a single burst spike. Finally, the closest matching configuration is selected based on the Euclidean distance metric defined as

\begin{equation*}
    \begin{split}
        cost = &(sic\_amplitude_{target} - sic\_amplitude_{config})^2 \\
        &- (sic\_window_{target} - sic\_window_{config})^2.
    \end{split}
\end{equation*}

Users can define their own range of values for the weights or current decays by regenerating the configurations table using the $sic\_data\_table.py$ file.

\section{Astrocytes induce Neuronal Synchronization}

\begin{figure}
\includegraphics[scale=1]{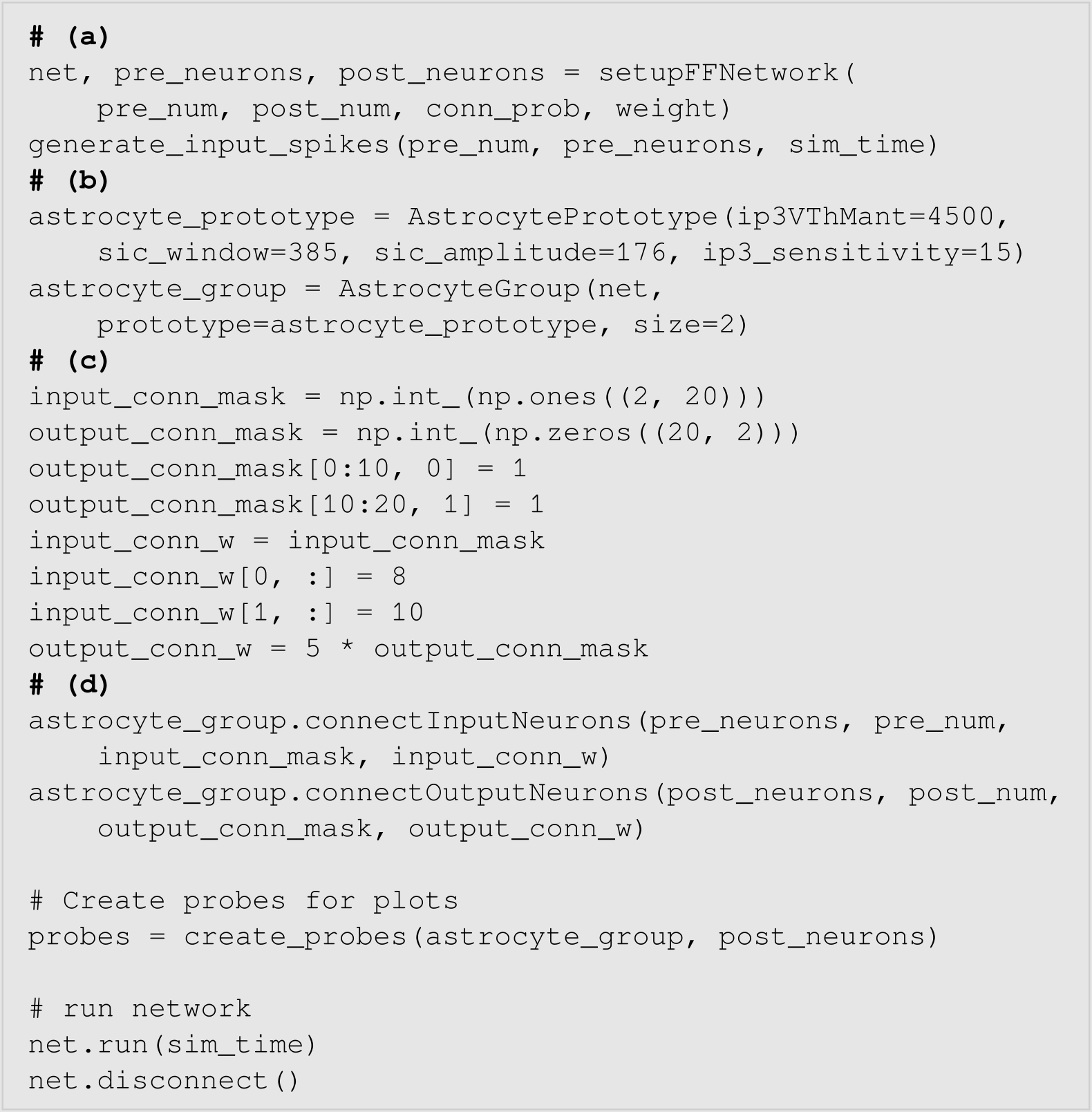}
\caption{An example Astrocyte SDK program for integrating 2 astrocytes into an SNAN using the AstrocyteGroup.}
\label{fig:example_astrocyte_program}
\end{figure}

\begin{figure}
\includegraphics[scale=1]{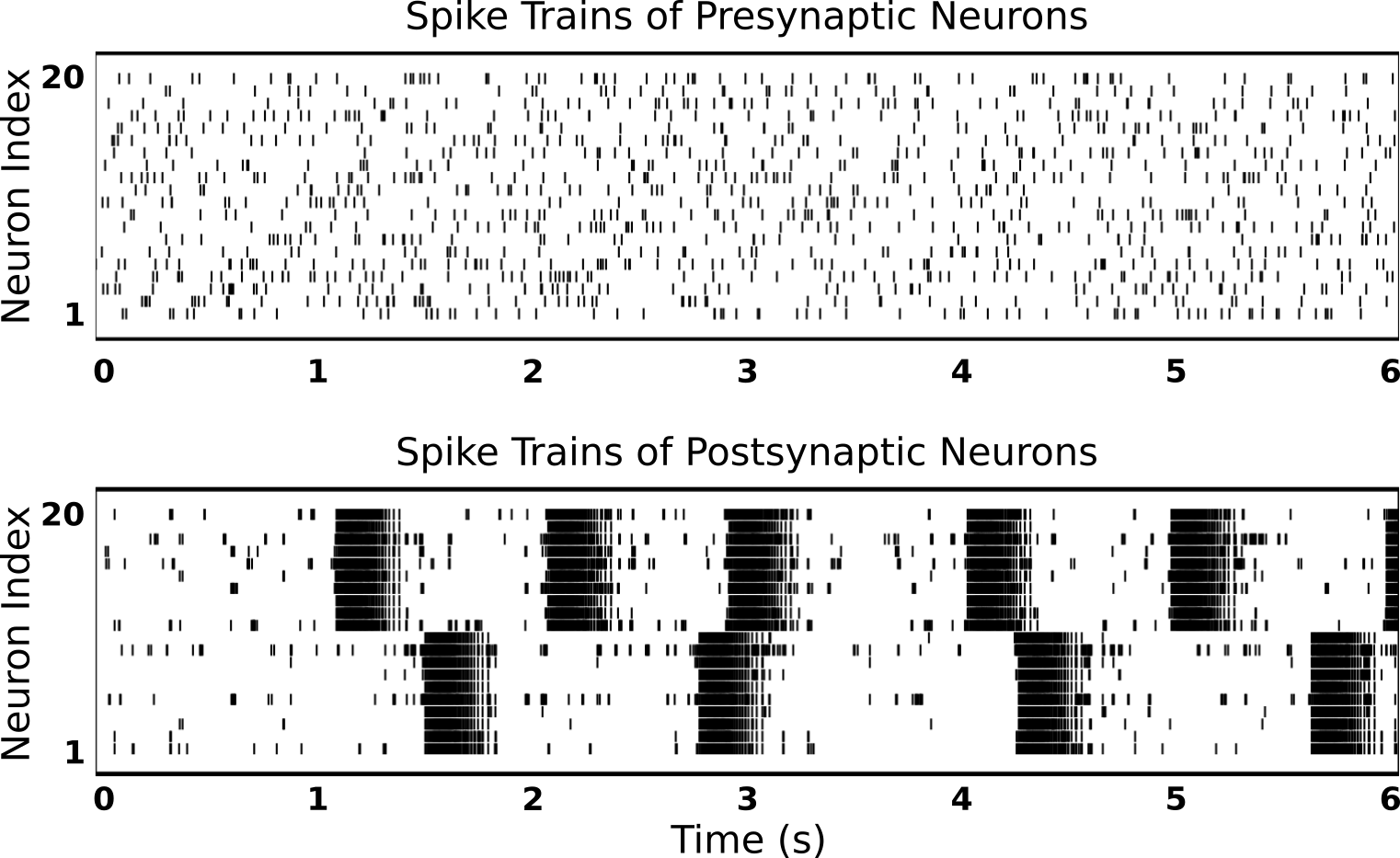}
\caption{An astrocyte group inducing synchronized activities in postsynaptic neurons.}
\end{figure}

We demonstrate how a group of astrocytes can be instantiated automatically, as described in Section 3.2, to synchronize sub-networks of neurons in an SNAN. We provide the code to create a single-layer feedforward SNN with 20 input neurons and 20 output neurons (Fig. 4). In this example, the synapses between the input and output neurons were randomly created with a probability of 8\% and a weight of 3. Briefly, to form the SNAN, we created the astrocyte group by defining an AstrocytePrototype (Fig. \ref{fig:example_astrocyte_program}a). We then created a group of astrocytes (Fig. \ref{fig:example_astrocyte_program}b) by specifying the NxNet instance it belonged to, the prototype(s) to use for instantiation and the number of astrocytes to create within the AstrocyteGroup. We also demonstrate how to form tripartite synapses based on the connection masks, indicating which of the possible neurons to connect to, and the weights for those connections (Figs. \ref{fig:example_astrocyte_program}c, d). Replicating experimental studies as well as our computational results \cite{Fellin, Knoxville}, the two Loihi astrocytes imposed distinct synchronization patterns in the neuronal component, forming two synchronized groups (Fig. 5).

\section{Astrocytes expand Hebbian Learning in Space and Time}

\subsection{Experimental and Computational Evidence}

Astrocytes are now known to take active part in several brain functions, from improving cognitive tasks and memory \cite{han2013} to suppressing futile unsuccessful behavior \cite{zebra}; Interestingly enough, they have also been found to not only be necessary but also sufficient for new memory formation \cite{adamsky2018astrocytic}. Astrocytes modify behavior by changing the network's structure in ways that cannot be captured by Hebbian-type rules. In fact, what distinguishes astrocytes from neurons is that they integrate synaptic activity over a much wider spatial scale and much longer temporal scale \cite{araque2014gliotransmitters}. This allowed us to explore possible mechanisms that astrocytes may use to increase or decrease synaptic weights, individually from neurons. 

The newly identified mechanisms in astrocytes can be used to expand two main characteristics of Hebbian-type learning. First, astrocytes can be used to introduce regional learning: Hebbian learning is localized and lacks any knowledge of the neuronal activity in the vicinity of the synapse. Any change in synaptic weights is driven by the spiking activity of the pre- and post-synaptic terminals that form a single synapse. However, the spatial scale of astrocyte-induced plasticity spans from neighboring spines \cite{Zhang} to synapses found hundreds of micrometers away from the active synapse \cite{purinergic}, extending the spatial reach of plasticity. Second, astrocytes can integrate synaptic activity over time, ranging from hundreds of milliseconds to seconds, extending plasticity beyond the short-term correlations between pre- and post-synaptic activities. Therefore, a change of synaptic weights that currently depends only on very short temporal dependencies, can now be influenced by the activity in a much larger time window, allowing, e.g., for temporally delayed plasticity effects \cite{Serrano, Letellier}. Such computational primitives step away from the typical Hebbian learning rules and can endow astrocytes with unique computational and learning abilities that are applicable to SNN.

\subsection{Astrocytic-induced Heterosynaptic Depression (HSD)}

Heterosynaptic plasticity offers a straightforward astrocyte - implicated extension of Hebbian learning, in time and space. For example, astrocytes are found to detect the high (tetanic) activity of a pathway and respond to it by inducing presynaptic depression on the neighboring inactive pathways \cite{Serrano}. This astrocyte-driven Heterosynaptic Depression (HSD) may be used to complement the associative nature of STDP by counterbalancing the homosynaptic alterations, making the heterosynaptic modifications more prominent. In the next session, we show how HSD can use its spatial information to augment STDP, introducing single-shot memories.

\begin{figure*}
\includegraphics[scale=1]{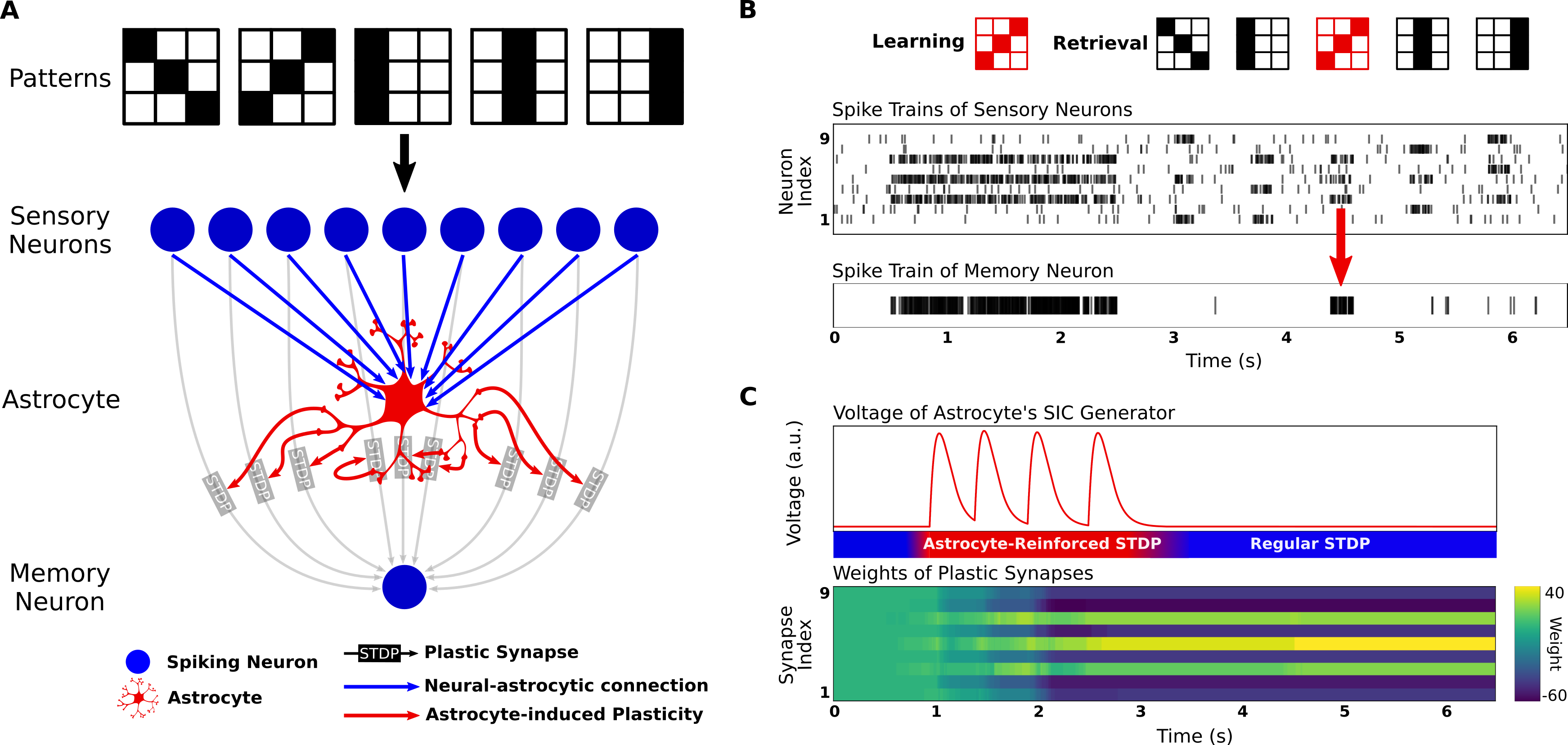}
\caption{Single-shot pattern memorization and retrieval. A. Architecture of the SNAN with astrocyte-reinforced STDP. B. Activity of the sensory neurons and the memory neuron during learning and retrieval. C. Dynamics of the synaptic weights potentiated by the astrocyte: Switching STDP on and off is based on the temporal persistence of the learnable input.}
\end{figure*}

\subsection{Example Implementation of HSD in LAM to Introduce Single-Shot Memory}

We now describe how LAM supports astrocyte-driven plasticity mechanisms, to incorporate space and time in synaptic weight modification. To showcase this functionality, we created a 2-layer feedforward SNAN (Fig. 6A), with the astrocyte monitoring the activity of the sensory neurons. LAM uses the reward channels of Loihi's learning engine to implement the astrocyte-induced HSD on top of the regular neuronal STDP component. 

The astrocyte-reinforced learning rule is defined as a sum of trace products \cite{lin2018programming},
\begin{equation}
    dw = \sum_i S_i \prod_j T_{ij},
\end{equation}
where $T_{ij}$ is the trace and $S_i$ is the learning rate. 

The regular STDP component implemented on Loihi is defined as
\begin{equation}
    dw_{stdp} = a*x_1*y_0-b*x_0*y_1,
\end{equation}
where $x_0$ and $y_0$ represent the presence of pre- and post-synaptic spikes, respectively, and $x_1$ and $y_1$ are the pre- and post-synaptic spike traces, decaying exponentially. 

The astrocyte-induced HSD component is defined as
\begin{equation}
    \begin{split}
        dw_{hsd} & = -(c*y_0 - d*x_0)*r_1 \\
        & = -c*y_0*r_1 + d*x_0*r_1,
    \end{split}
\end{equation}
where the reward spike trace $r_1$ represents the activity of the astrocyte. The HSD component decreased the weights of all neuronal synapses when the astrocyte was active. For more prominent heterosynaptic changes, the component induced slower weight decreases for sensory neurons with stronger activity and faster decreases for those with weaker activity.

We finally define the HSD-reinforced STDP on Loihi as
\begin{equation}
    dw = dw_{stdp} + dw_{hsd}
\end{equation}

\subsubsection{Experiment}

We demonstrate how LAM can be used to enhance the regular STDP learning rule by superimposing an HSD-reinforced component in a feedforward SNAN (Fig. 6A). The network consisted of 9 sensory neurons, each of which encoded one block of a 3x3 input grid into a Poisson spike train. The input neurons were connected to one output (memory) neuron. An astrocyte monitored the synaptic connections and modulated the memory neuron by injecting a SIC. We used 5 input patterns, with each pattern having 3 neurons active. All patterns had an overlap of one active neuron. The input neurons had a baseline firing rate of 5 Hz (white blocks) and an active firing rate of 100 Hz (black blocks). The Loihi parameters used in this implementation were as follows: $a=2^{-5}$, $b=2^{-6}$, $c=2^{-2}$, $d=2^{-1}$, the impulses for $x_1$ and $y_1$ were 16 whereas for $r_1$ was 8, and the decay time constants for all traces were 2. First, the network was trained on a desired pattern, which was presented for 2 seconds (memory learning). Then, the different overlapping patterns were presented to the network for 0.2 seconds each (memory retrieval). 

\subsubsection{Results}
By virtue of its own slow dynamics, the astrocyte switched on and off STDP learning, controlling when the network memorized the pattern. Specifically, when an input pattern was presented to the network long enough (in the order of seconds), this persistent neuronal activity triggered the astrocytic SIC. Further, as it integrated inputs from regional synapses, the astrocyte became active only when the overall input was strong enough. When the astrocyte was activated, all synapses encoding inactive blocks decreased their weights to negative values through HSD, thereby inhibiting the memory neuron (Fig. 6C). This created a distinction between regular STDP and the astrocyte-reinforced STDP: With regular STDP, any active block that overlaps among patterns (e.g. the central block in the first two and the fourth patterns - Fig. 6A) increases the weight of its synapse every time a pattern is presented. In the presence of HSD, when active blocks overlapped across patterns, the memory neuron received inhibitory input from the sensory neurons corresponding to active grid blocks forming a pattern different than the one learned. During memory retrieval, the memory neuron fired maximally when the learned pattern was presented (Fig. 6B). Neurons that were irrelevant to the learned pattern had negative weights and inhibited postsynaptic firing. With no astrocytic contribution, this learning rule downgraded to a regular STDP (Fig. 6C) that does not change the synaptic weights when the postsynaptic activity is low. In other words, the astrocyte controlled \textit{when} the network memorized a new pattern, based on the input's persistence. Overall, LAM enables an astrocytic gating mechanism for neuronal learning that expands the regular STDP to space and time.

\section{Astrocytes Learn to Monitor Transition into Chaos}

\subsection{LAM Support of Astrocytic Learning}


LAM enables the creation of astrocytes with custom input plasticity mechanisms that are limited only by the set of operations supported by Loihi. It is known that astrocytes perform large-scale spatial and temporal integration of synaptic activity \cite{perea2009tripartite,araque2014gliotransmitters} and exhibit bidirectional homeostatic plasticity (BHP), where astrocytic connections continuously increase/decrease sensitivity to synapses with low/high activity over the course of several hours \cite{xie2012}. We incorporated rate-dependent astrocytic learning in LAM as follows:
\begin{equation}
    \begin{split}
        dt &= f_{r}(x_0)\\
        dw &= f_{w}(w,t),
    \end{split}
\end{equation}
where $f_{r}$ dynamically integrated synaptic input spikes sensed by the astrocyte, resulting in a spike rate approximation variable $t$ for each tripartite synapse, and $f_{w}(w,t)$ implemented astrocytic weight learning defined in the astrocyte input compartment. In the next session, we demonstrate how the proposed astrocytic model with rate-based learning may detect the transition of the neuronal network, from order to chaos.

\subsection{An Astrocytic Model For Detecting Network Chaos}

SNNs fall under the general category of complex nonlinear dynamical systems, which maximize their computational capacity when they operate at the edge of chaos, as shown by studies ranging from cellular automata to boolean networks \cite{packard1988adaptation,ribeiro2008}. It is at the critical point between ordered and chaotic dynamics where balance of robustness and versatility emerges not only in brain-imitating networks \cite{legenstein2007edge} but also in the brain itself \cite{friedman2012universal}. Interestingly, the design of learning algorithms largely does not take into account this general computing principle that would enable SNNs to operate at the edge of chaos. In our ongoing effort to develop self-tuning near-critical SNNs, we proposed and developed a homeostatically plastic, astrocyte model capable of detecting and signaling the approach of network chaos, detailed in our modeling paper \cite{vlad2019unpublished}, and abstracted here as,
\begin{equation}
f_{astro}(r_{i}, {\hat{r}}_{i}) = g \Bigg( \bigg( \frac{1}{N} \bigg) \sum_{i=1}^{N} \frac{r_{i}}{r_{max}} \log{\Bigg(   \frac{1}{\frac{{\hat{r}}_{i}}{r_{max}}}\Bigg)} \Bigg),
\end{equation}
where  $r_i$ is the short-term synaptic firing rate, ${\hat{r}}_{i}$ is the long-term synaptic firing rate, $r_{max}$ is the maximum synaptic firing rate, $g$ is a nonlinear activation function, and $N$ is the number of synaptic inputs into the model.  Intuitively, this model compares the run-time, short-term synaptic firing rates, $r_i$, to the memorized, long-term synaptic firing rates, ${\hat{r}}_{i}$, and aggregates the extent of the individual synaptic differences in the overall astrocytic output frequency, $f_{astro}$.

\begin{figure*}
\includegraphics[scale=1]{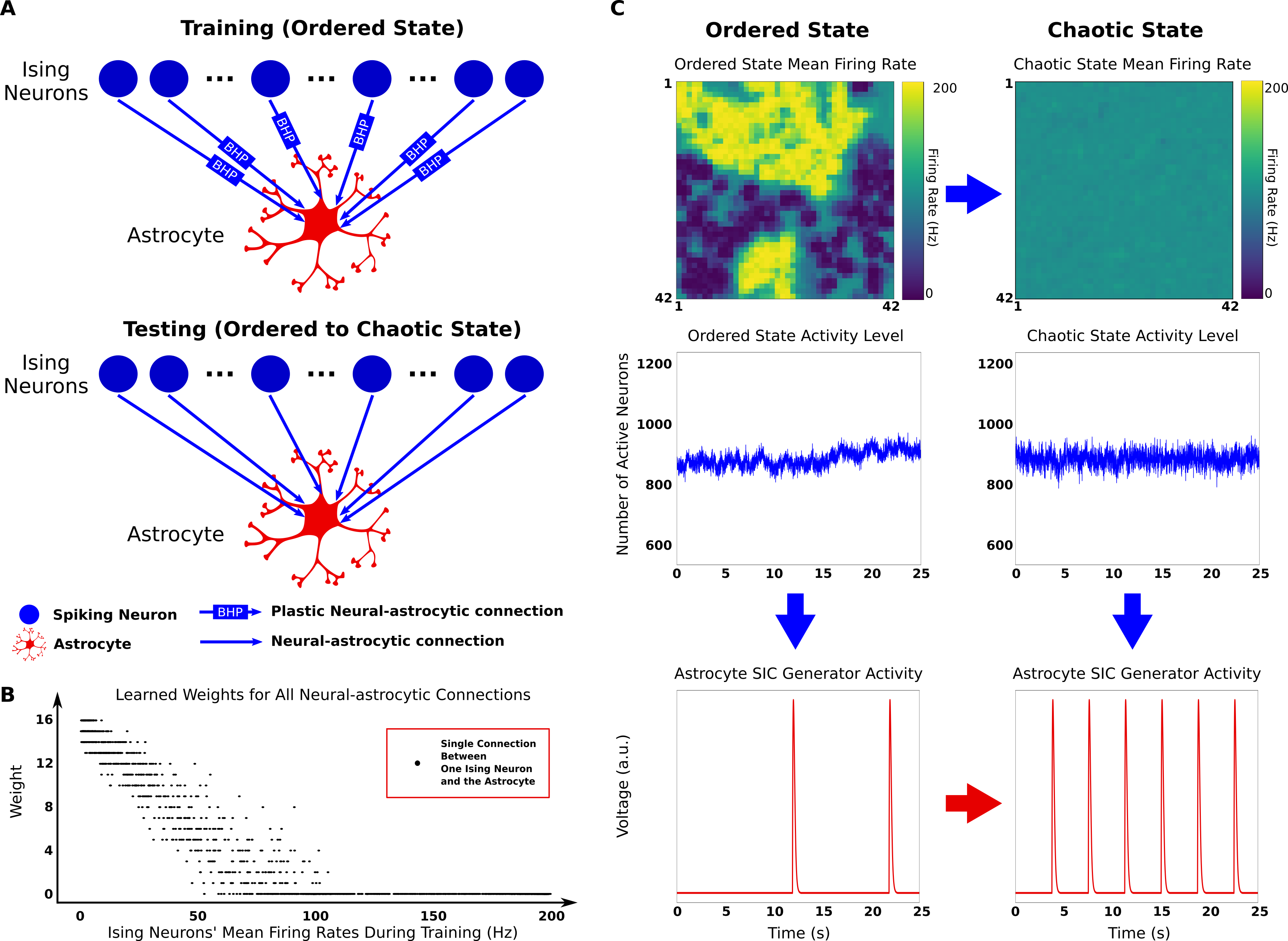}
\caption{Homeostatic astrocyte on Loihi sensing transition from ordered to chaotic network state. A. SNAN architecture showing training and testing phases. B. Learned weights for all the neuronal-astrocytic connections. C. (upper row) Ordered and chaotic neuronal network state (middle row) number of active neurons in both states showing similar neuronal activity (bottom row) astrocytic activity during the transition from order to chaos.}
\end{figure*}

\subsection{Implementation of Astrocytic Learning in LAM to Detect Chaos}

We implemented an SNAN with a single-layer of neurons connected to an astrocyte endowed with the BHP learning mechanism (Fig. 7A). To approximate the BHP rule on Loihi, we decomposed Equation (5) into a weighted sum,
\begin{equation}
    \begin{split}
        f_{astro}(r_{i}, {\hat{r}}_{i}) = g \Bigg(\eta \sum_{i=1}^{N} W_i*N_i\Bigg) \\
        W_i  = \log{\Bigg(   \frac{1}{\frac{{\hat{r}}_{i}}{r_{max}}}\Bigg)} ,\ N_i  = \frac{r_{i}}{r_{max}}.
    \end{split}
\end{equation}

The input $N_i$ was represented as the integrated current of real-time presynaptic spikes $I_i$ and the parameter $W_i$ was the weight of the neural-astrocytic connection $w_i$, described by
\begin{equation}
Activity_{astro} = Astrocyte(\sum_{i=1}^{N}w_i*I_i).
\end{equation}

Following the general form in Equation (5), the weight $w_i$ was learned based on the BHP learning rule, defined as,
\begin{equation}
    \begin{split}
        dt &= a*x_0\\
        dw &= b*u_k*(w_{max}-w)-c*u_k*t,
    \end{split}
\end{equation}
where $t$ represents the estimated long-term presynaptic neuron firing rate, and $w$ approximates the inverse relationship between the neuronal-astrocytic weight and the presynaptic firing rate described by $W_i$. Intuitively, the learning rule kept decreasing/increasing $w$ for high/low $t$ values in a predefined learning time window, where $t$ did not saturate (Fig 7B.). 

\subsubsection{Experiment}

The astrocyte was trained using the BHP rule (Eq. (8)) on a single layer of spiking neurons driven by activity that was generated off-chip. To represent the neuronal activity, we used a non-isotropic Ising model, which is described in \cite{vlad2019unpublished}. Briefly, we modeled synaptic activity as 256x256 magnetic spins positioned on a 2-dimensional lattice with non-uniform, clustered couplings between spins resulting in stationary spin state patterns. We used the Markov Chain Monte Carlo (MCMC) algorithm to simulate the evolution of spin states by minimizing the system's overall energy. We determined ordered and chaotic Ising states (Fig. 7C) using the magnetic susceptibility measure \cite{yeomans1992statistical}. To drive the neurons on Loihi, we downsampled 42x42 spins from the original Ising system, evenly distributed throughout the Ising lattice, and converted each binary Ising spin state to a corresponding neuronal spike state. All Ising spins were updated and transformed into neuronal spikes every 5 milliseconds, keeping the maximum neuron firing rate to 200 Hz (Fig. 7C). The astrocyte's receptor weights were trained for 25 seconds on neuronal dynamics driven by ordered Ising spin evolution. Then, the model's activity was tested with respect to both states for 25 seconds each. The parameters for the BHP rule were $a=2^{-6}$, $b=2^{-2}$, $c=2^{-3}$, $w_{max}=16$, and $k=4$ in $u_k$ which controlled the $w$ to only update every $2^4$ learning epochs.

\subsubsection{Results}

The frequency of the astrocytic calcium wave increased when the neuronal component transitioned from an ordered to a chaotic state of neuronal firing (Fig. 7C). This nearly 200\% increase in frequency resulted from the weights that were learned through the BHP rule, since network activity levels remained unchanged throughout the transition (middle row, Fig. 7C). Specifically, the learned weights generated by the BHP inversely correlated with the long-term mean firing rates of the Ising neurons during training (Fig. 7B), imitating the functional form of $W_i$ in Equation (6). As the minimally active neurons increased their firing rates with the transition to chaos (first row, Fig. 7C), their higher weights induced increased input current to the astrocyte, which resulted to increased astrocyte activity. That is how the transition between states was monitored by a single astrocytic cell. 

\section{Discussion and Conclusion}

Here, we introduced LAM, a module that enables the integration of astrocytes to Loihi, endowing to SNN the computational primitives recently identified in these long-neglected non-neuronal cells. Adding astrocytes to neuromorphic processors, which are built on the premise that the neuron is the \textit{only} computational unit in the brain, has a great potential to advance neuromorphic computing. To demonstrate this potential, we demonstrated three example uses of astrocytic computation, based on several pathways for astrocyte-neuron interaction and neuromodulation.

Following biological evidence that astrocytes exhibit homosynaptic \cite{Perea2014} and heterosynaptic \cite{Serrano} plasticity mechanisms, at both presynaptic \cite{prePlast} and postsynaptic \cite{postPlast2} sites, LAM supports (i) HSD, shown here to be implicated in single shot pattern memorization, and (ii) BHP, demonstrated as they key mechanism for monitoring the transition of the neuronal network to chaos. These astrocytic learning primitives showcased how network learning can extend beyond updating the weights of the neuronal synapses and onto the wider operating time and space of astrocytes.

In addition, the astrocytes in LAM were used to induce neuronal synchronization, replicating both experimental and computational studies \cite{Knoxville,tewari2013possible}. Astrocytes are long-implicated in network frequency modulation \cite{tewari2013possible,LeoRhythms}, improving network polychronicity \cite{valenza2013novel} and enabling memory transition in attractor networks \cite{LeoMemory}. In our example, time and space became learnable parameters beyond the reach of Hebbian-type learning, allowing the introduction of “on-demand” synchronization or binding, principles in the brain that are long-believed to process information efficiently but are missing in asynchronous neuromorphic hardware.

Finally, we showed here how astrocytes in LAM may induce heterosynaptic depression constrained within their microdomain. By introducing spatial processing, astrocytes can further induce local plasticity and, through it, synaptic clustering, which is a computational primitive for robust brain computation that is also missing in neuromorphic solutions. 

In taking our next steps to expand LAM, it has not escaped our attention that neuromorphic chips may be used as a computational framework enabling astrocytic functions to be both explored in brain hypotheses and exploited for computing purposes. Currently, LAM develops astrocytes as dimensionless units, having no inter-unit communication. A natural extension for our module is to extend astrocytic computation to (i) the subcellular level and (ii) network level. Both imaging \cite{DiCastro} and stimulation techniques \cite{Zhiping} provide evidence that the majority of astrocytic activity is spatially confined to functionally independent subcellular astrocytic compartments, rarely spanning the whole cellular domain. Adding dimensions in our astrocytic cells and networks would enable them to operate, process and learn, on a semi-local, "regional" or long-range scale. In fact, astrocytes, coupled through inter-astrocytic gap junctions, form a network syncytium, thereby enabling long distance communication on top of neuronal networks \cite{Bell}. Enabling both multi-compartment processing and inter-astrocytic communication would further bridge the computation capabilities of LAM with those of biological astrocytes. 

Overall, although they probably require a lot more small insights before they can be fully integrated to SNN, the fact that our Loihi-run SNAN already emulate biological principles associated with robustness and versatility, indicates that the integration of astrocytes to neuromorphic chips is a direction worth exploring.

\begin{acks}
This research is funded by Intel's NRC Research Award.
\end{acks}

\bibliographystyle{ACM-Reference-Format}
\bibliography{sample-bibliography}

\end{document}